\documentclass{article}
\usepackage[dvips]{graphicx}
\usepackage[utf8]{inputenc}
\usepackage{amssymb,amsmath,array}

\usepackage{todonotes}
\usepackage{caption}
\usepackage{subcaption}
\usepackage[capitalize]{cleveref}
\usepackage{url}
\newcommand{\D}{\mathcal{D}}

\begin{document}
\title{Does Better Drift Detection Always Yield Better Models? An Empirical Study on Stream Learning}

\author{Valerie Vaquet$^\dagger$ and Fabian Hinder$^\dagger$
%
\thanks{Funding in the scope of the BMFTR project KI Akademie OWL under grant agreement
No. 16IS24057A is gratefully
acknowledged.}
%
\vspace{.3cm}\\
%
Bielefeld University - Faculty of Technology \\
Inspiration 1, 33619 Bielefeld – Germany\\
$\dagger$ Authors contributed equally
%
}

\date{}

\maketitle

\begin{abstract}
\noindent Stream learning algorithms often rely on drift detectors to trigger model updates, implicitly assuming that accurate detection ensures better predictive performance. However, the interplay between detector quality and model accuracy remains underexplored. This work empirically investigates this relationship using synthetic datasets with real drift. We evaluate four hypotheses regarding the impact of correct detections, false alarms, detection delay, and repeated updates on model error. 
Our results reveal that drift detection is beneficial in scenarios with severe drift. In milder drift settings, false alarms and excessive updates can degrade performance. We conclude that active adaptation strategies must be carefully calibrated to the specific drift characteristics, as passive adaptation can be equally effective when drift is mild.
\end{abstract}

\section{Introduction}
Real-world data is frequently collected over time and potentially affected by changes in the underlying data distribution -- a phenomenon formalized as concept drift, or drift for short \cite{gama_survey_2014}. Drift can be caused by seasonal changes, sensor or hardware degradation or failure, etc. Since concept drift contradicts the assumptions of batch learning (the data being iid), in this setup, we consider stream learning models aiming to keep high predictive performance despite the drift \cite{gama_survey_2014}. 

To adapt to the current concept well, many stream learning algorithms rely on a drift detector to trigger model updates \cite{ditzler_learning_2015}.
Here, the implicit assumption is that a well-performing drift detector, which accurately detects drifts with a negligible delay, leads to a well-performing model.
However, when evaluating such stream learning algorithms, usually the predictive error of the entire model is the primary focus. While some works also report detection metrics, a precise evaluation of the interrelation of detector quality and model performance is rarely applied, thereby preventing a comprehensive understanding of their interplay. Thus, in this work we will investigate the question: What is the relationship between the performance of the drift detector and the predictive model?

In particular, we check the following hypotheses motivated by implicit assumptions made in many works:
\begin{itemize}
    \item[H1] \emph{Detecting drift benefits the stream learning performance.} Detection can trigger appropriate adaptation in stream learners if a drift occurs, thereby avoiding performance degradation. 
    \item[H2] \emph{Many false alarms diminish model performance.} False alarms can trigger unnecessary updates, thereby disrupting the learning process.
    \item[H3] \emph{Timely detection is important.} Detecting drift without delay can trigger model updates immediately, thereby avoiding severe performance drops. 
    \item[H4] \emph{Too many correct alarms are harming model performance.} Repeated updates triggered by alarms disturb the updating procedure even when caused by a ground truth drift.
\end{itemize}

\section{Foundations and Related Work}
A data stream is formalized as an infinite sequence of independent observations $(X_1,Y_1), (X_2,Y_2), \dots$, each drawn from a potentially different distribution $(X_i, Y_i) \sim \D_i$, with drift referring to $\D_i\neq \D_j$ for some $i, j$. Usually, drift is further categorized into \emph{real drift}, referring to changes in $\D_i(Y\mid X)$, and \emph{virtual drift}, referring to changes in $\D_i(X)$ that do not affect $\D_i(Y\mid X)$ \cite{gama_survey_2014,hinder_hardness_2023}.

In such settings, we consider two primary tasks: \emph{Drift detection} aims at detecting drifts based on the observations. \emph{Stream learning} focuses on adaptive models that perform well on the most recent concept by employing suitable update strategies when faced with drift \cite{ditzler_learning_2015}. For the latter, there are two main strategies: While \emph{passive adaptation} schemes constantly update regardless of drift, e.g., by retraining on sliding windows, \emph{active adaptation} strategies additionally employ a drift detector to force retraining when drift is detected \cite{ditzler_learning_2015}. Many works exploring active strategies follow the intuition that drift harms the model performance and that, hence, a drift detector should be used to trigger action. Following this argument, a well-performing drift detector should be correlated with a well-performing active stream learner. In this work, we will investigate this intuition. Before performing our experimental evaluation, we recap the most important related results.

\cite{hinder_hardness_2023} investigated drift detection with a focus on spotting drift for system monitoring, i.e., with the goal of analysing and understanding the data without a focus on automated learning. They found that real and virtual drift are not distinguishable due to limited model complexities in practical applications. Furthermore, \cite{hinder_change_2023} analyzed whether changes in model loss imply actionability for stream-learning models. They found that acting by updating the model is only beneficial if the theoretical optimal model loss is constant. Consider, for example, that a drift decreases data quality, for instance, by inducing more label noise, while the optimal decision boundary, i.e., the theoretical optimal model, stays constant but the optimal loss decreases. In such a setting, the information a stream learning model acquired is still valid. Thus, an update could harm model performance even though the drift detector correctly detected drift.

A related practical evaluation \cite{hinder_real_2026} investigated the question of how probable an increase in model loss is across different settings, involving both synthetic and real-world datasets with real and virtual drift, respectively. They found that the loss only marginally increases on the evaluated real-world datasets. Besides, they could not confirm the folklore that real drift is more harmful for model performance than virtual drift for the considered real-world datasets, yet found that pattern in the considered synthetic data streams. \cite{hinder_real_2026} focused on the impact of drift on model loss but did not consider the role of the drift detector, which we will investigate in this work.

\section{Experimental Study}
To answer our research questions empirically, we run an experimental study.\footnote{Code available at \url{https://github.com/FabianHinder/Better-Detection-Better-Models}} We considered several popular synthetic benchmarks (Agrawal, RandomRBF, SEA, and Sine; all from River). For each stream type, we generated 500 streams of length 1000 with one randomly drawn abrupt drift event between 450 and 550. We decided to rely only on synthetic datasets with abrupt real drift, as those yielded the most pronounced increase in model loss in \cite{hinder_real_2026}. Lower ones would remain undetected, yielding very limited material to analyze. This study examines abrupt drifts as they are considered to be a suitable setting for active approaches relying on drift detection \cite{ditzler_learning_2015}.

Model-wise, we evaluated a $k$NN classifier, logistic regression, and a Hoeffding tree, aiming to represent the most prominent model classes in stream learning, using River's default hyperparameters, up to standard scaling in the pipeline for $k$NN and logistic regression, and reduced the $k$NN window size to 200 to match the lengths of the considered data streams. All stream learners are passively adapting throughout the stream; additionally, in case the drift detector alerts, the model is reset. Regarding drift detectors, we considered FHDDM \cite{fhddm}, ADWIN \cite{adwin}, and KSWIN \cite{kswin}, covering popular choices again using default hyperparameters up to considering all detection threshold choices $\alpha \in [0.05,0.01,0.005,0.001,0.0005,0.0001]$. As baselines, we consider an oracle drift detector (Orac.), alerting drift events correctly, and no detection, leading to a fully passive approach (Pass.).

For the evaluation, we decided to consider an alarm as a true positive (TP) given it occurred within 100 samples of the ground truth drift. Note that here we evaluated a symmetric case where an alarm happening up to 100 samples before the ground truth drift also counts as a TP, as resetting a model just before a drift occurs results in a similar model compared to resetting it shortly after the drift occurs. All alarms raised outside of this window are considered false positives (FP). Note that we also performed an ablation study using the more traditional non-symmetric window, counting alarms as TPs only after the drift point (see arXiv version). Overall, we only observed slight changes in the results, which do not alter the interpretation.

Based on this experimental setup, we evaluated our hypotheses and came to the following results. 

\begin{figure}[t]
    \centering
        \begin{subfigure}[b]{0.48\textwidth}
        \includegraphics[width=\textwidth]{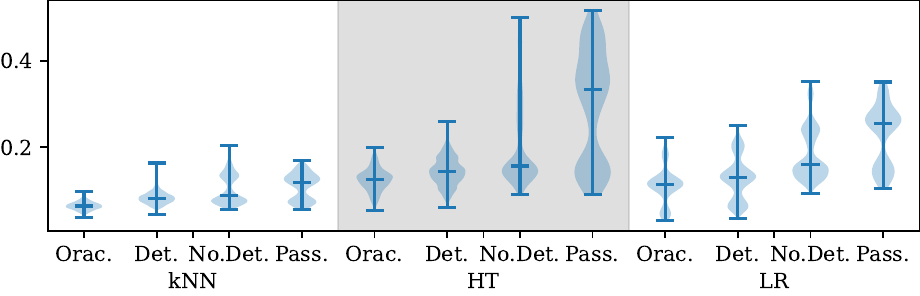}
        \caption{Sine}
        \label{fig:h1-sine}
    \end{subfigure}%
    \begin{subfigure}[b]{0.48\textwidth}
        \includegraphics[width=\textwidth]{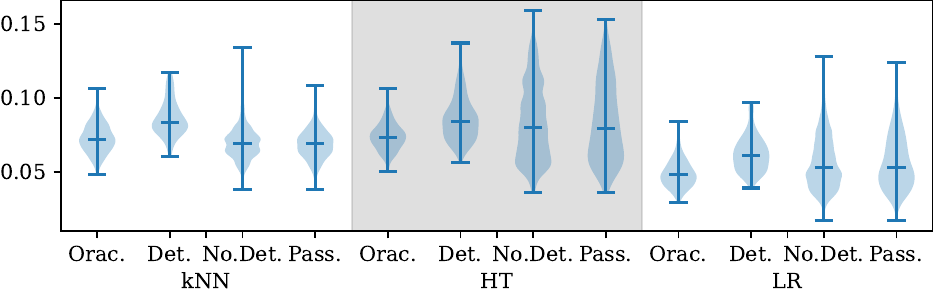}
        \caption{SEA}
        \label{fig:h1-sea}
    \end{subfigure}
    
    \begin{subfigure}[b]{0.48\textwidth}
        \includegraphics[width=\textwidth]{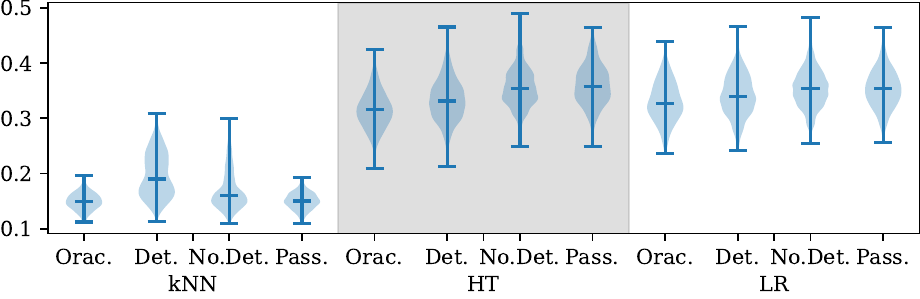}
        \caption{Random RBF}
    \end{subfigure}%
    \begin{subfigure}[b]{0.48\textwidth}
        \includegraphics[width=\textwidth]{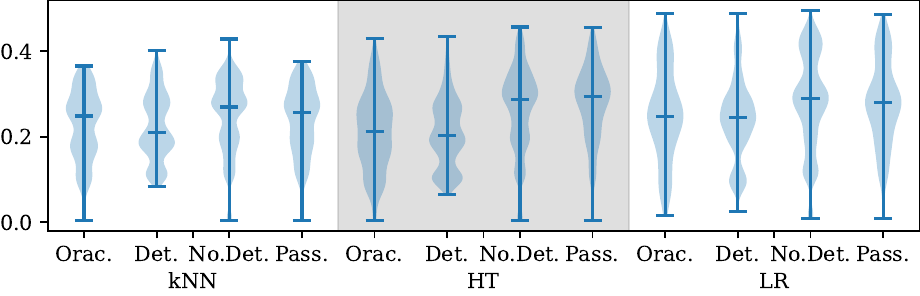}
        \caption{Agrawal}
    \end{subfigure}
    \caption{ITTEs, grouped by ground oracle drift detector (Orac.), streams where the detector detected at least one true positive drift (Det.) and where it did not (No Det.), and passive approaches (Pass.).}
    \label{fig:h1}
\end{figure}

\paragraph{Hypothesis 1:} Detecting drift benefits the stream learning performance\\
The ITTEs of our experiments are visualized in \cref{fig:h1}. While drift detection is clearly beneficial on the Sine dataset (\cref{fig:h1-sine}), we observe that on SEA (\cref{fig:h1-sea}) the settings where the drift detector raised a true positive alarm yield higher error rates than those without detection and the passive baseline, which perform approximately on par with the ground truth detection. For Agarwal, we observe lower ITTE for successful drift detection and the oracle baseline. In contrast, on the Random RBF dataset, the observations strongly depend on the classifier, with minor improvements for Hoeffding trees and logistic regression when applying drift detection successfully and contrasting observations for a $k$NN classifier.

Thus, while H1 holds for severe drift (e.g., Sine, high magnitude due to label flip), we cannot confirm it for mild drifts (e.g., SEA, low magnitude due to limited parameter shift). We will explore possible causes for this observation by analysing the effects of FPs, TPs, and detection delays. 

\begin{figure}[t]
    \centering
        \begin{subfigure}[t]{0.45\textwidth}
        \includegraphics[width=\textwidth]{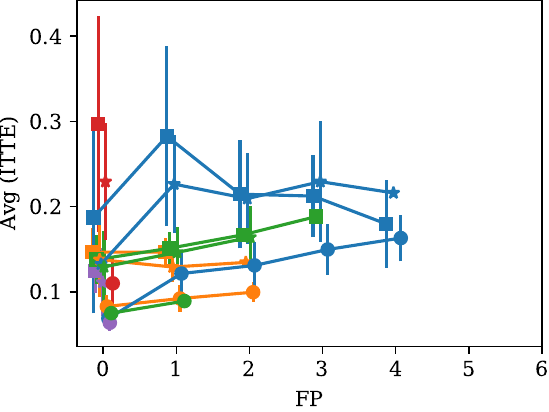}
        \caption{Sine}
        \label{fig:h2-sine}
    \end{subfigure}%
    \begin{subfigure}[t]{0.45\textwidth}
        \includegraphics[width=\textwidth]{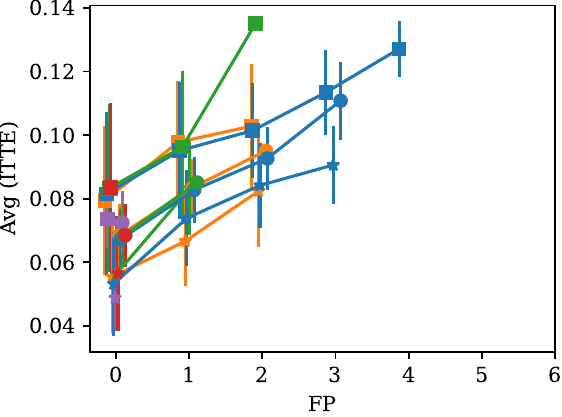}
        \caption{SEA}
        \label{fig:h2-sea}
    \end{subfigure}

    \begin{subfigure}[t]{0.45\textwidth}
        \includegraphics[width=\textwidth]{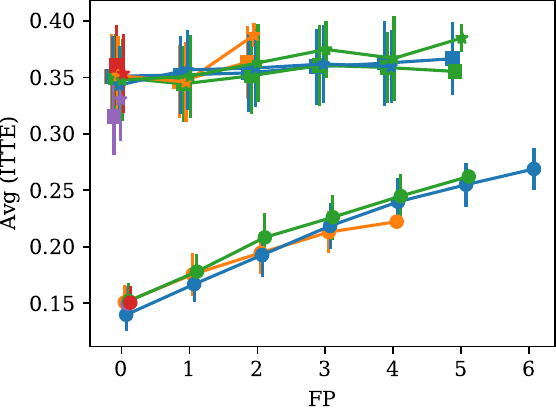}
        \caption{Random RBF}
        \label{fig:h2-rbf}
    \end{subfigure}%
    \begin{subfigure}[t]{0.45\textwidth}
        \includegraphics[width=\textwidth]{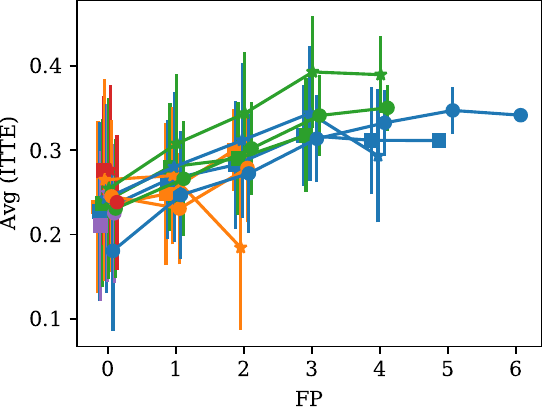}
        \caption{Agrawal}
        \label{fig:h2-agrawal}
    \end{subfigure}
    \begin{subfigure}[t]{\textwidth}
        \includegraphics[width=\textwidth]{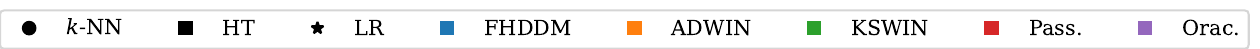}
    \end{subfigure}
    
    \caption{Average ITTE over increasing false positives for combinations of model and detection strategy}
    \label{fig:h2}
\end{figure}

\paragraph{Hypothesis 2:} Many false alarms diminish model performance\\
To test whether many false alarms increase the ITTE, we summarized our experimental results by the number of FPs; see \cref{fig:h2}. 
For most experiments, the reported ITTE increases for higher numbers of FPs. While this pattern is visually very clear for SEA (\cref{fig:h2-sea}) and Agrawal (\cref{fig:h2-agrawal}), it is harder to spot for Sine (\cref{fig:h2-sine}), partly due to the compressed y-axis scale. For RandomRBF (\cref{fig:h2-rbf}), we observe slight increases for Hoeffding tree and logistic regression models. In contrast, for the configurations containing $k$NN, we report clear increases in ITTE.

\begin{figure}[t]
    \centering
    \includegraphics[width=0.8\linewidth]{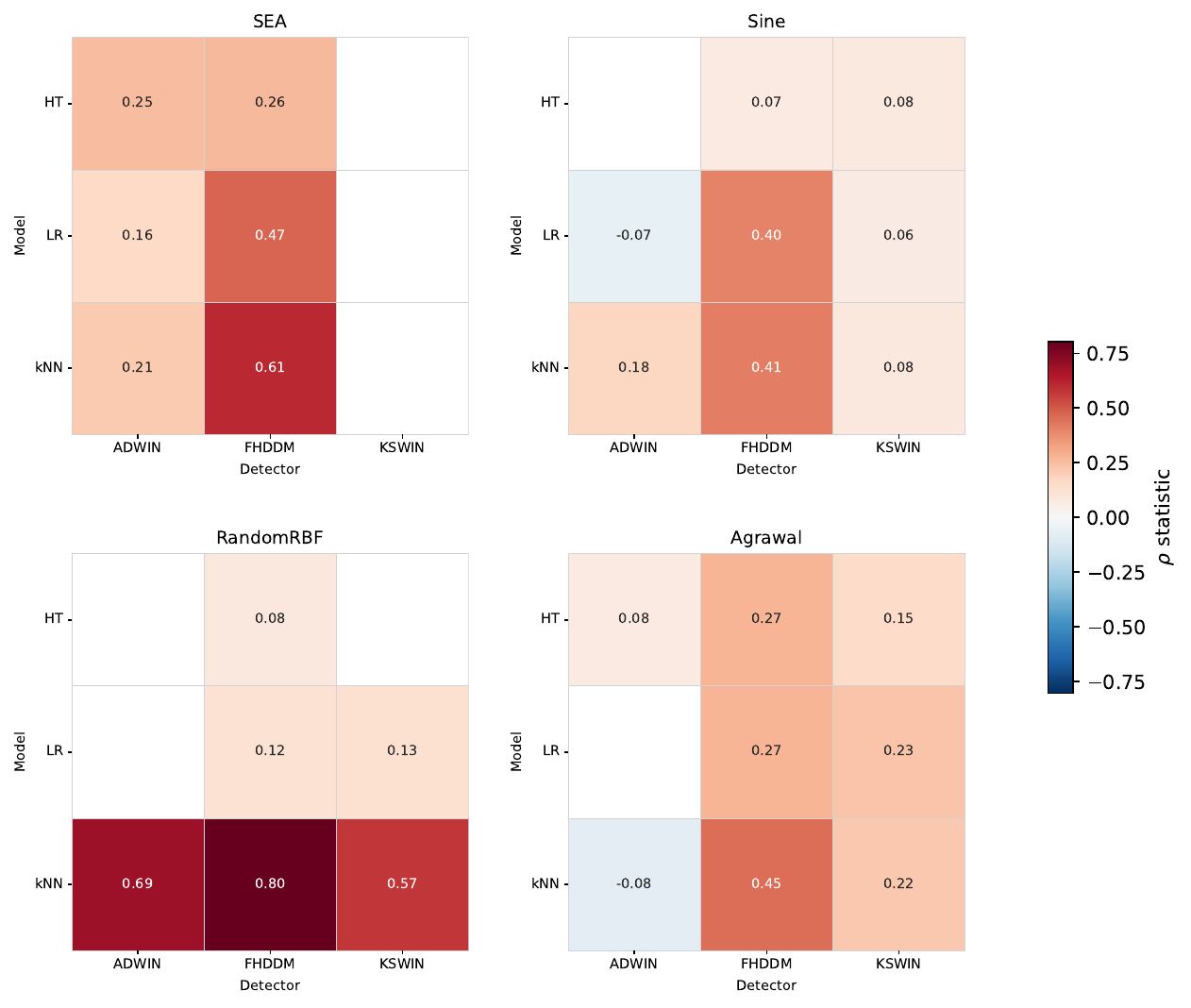}
    \caption{Results of Pearson correlation test; white fields indicate no statistically significant correlation, red indicates positive, blue indicates negative significant correlation}
    \label{fig:h2-stat}
\end{figure}
As shown in \cref{fig:h2-stat}, a Pearson correlation test ($p<0.001$) confirms the negative effect of many FPs on model performance, yielding a statistically significant positive correlation between FPs and ITTE for 26 configurations, and a negative for two -- both rely on ADWIN as the drift detector. The other 8 combinations did not yield significant results. 

Thus, we confirmed H2 in nearly all cases -- many false alarms harm model performance. This effect is more pronounced for datasets with lower drift magnitude, e.g., SEA; and more suitable model-dataset combinations, e.g., the combination of $k$NN \& RandomRBF, 
indicating that multiple resets hurt model performance more in scenarios with less severe drift. This also provides the primary explanation for H1.

\begin{figure}[t]
    \centering
        \begin{subfigure}[b]{0.48\textwidth}
        \includegraphics[width=\textwidth]{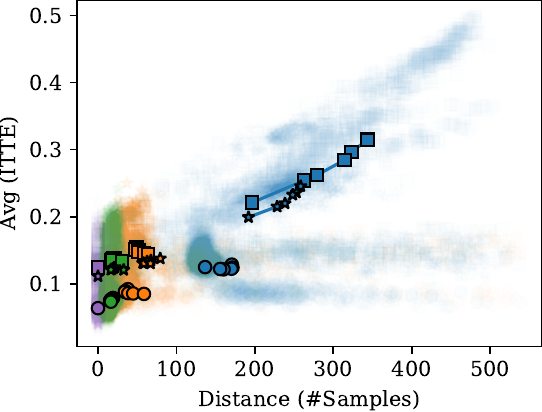}
        \caption{Sine}
        \label{fig:h3-sine}
    \end{subfigure}%
    \begin{subfigure}[b]{0.48\textwidth}
        \includegraphics[width=\textwidth]{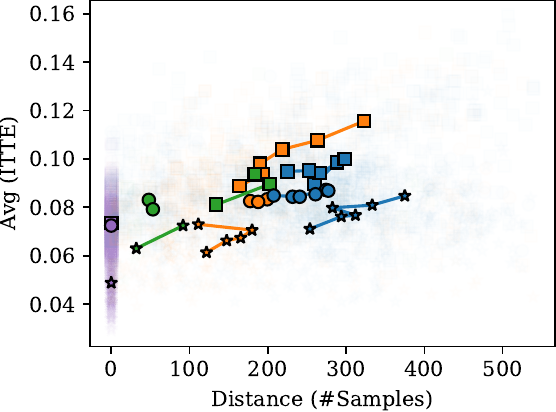}
        \caption{SEA}
        \label{fig:h3-sea}
    \end{subfigure}
    \begin{subfigure}[b]{0.48\textwidth}
        \includegraphics[width=\textwidth]{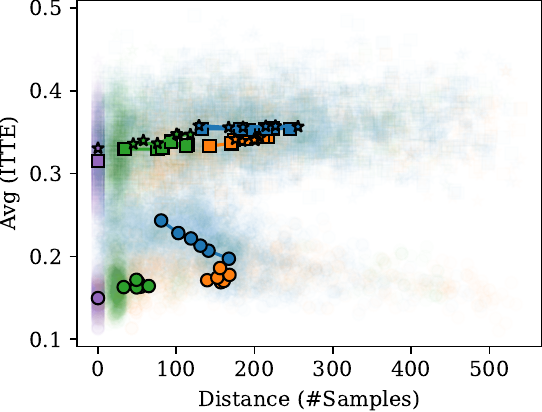}
        \caption{RandomRBF}
        \label{fig:h3-randomrbf}
    \end{subfigure}%
    \begin{subfigure}[b]{0.48\textwidth}
        \includegraphics[width=\textwidth]{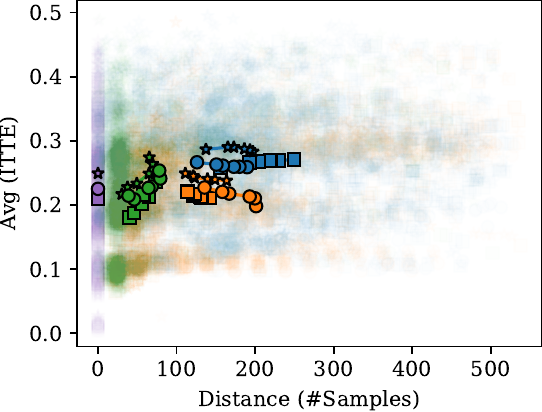}
        \caption{Agrawal}
        \label{fig:h3-agrawal}
    \end{subfigure}
    \begin{subfigure}[t]{\textwidth}
        \includegraphics[width=\textwidth]{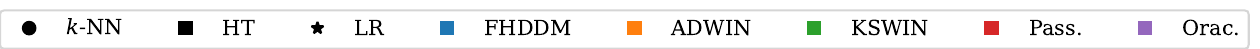}
    \end{subfigure}
    \caption{Average ITTE for varying distance of the first alarm for different combinations of model and detection strategy}
    \label{fig:h3}
\end{figure}
\begin{figure}[t]
    \centering
    \includegraphics[width=0.8\linewidth]{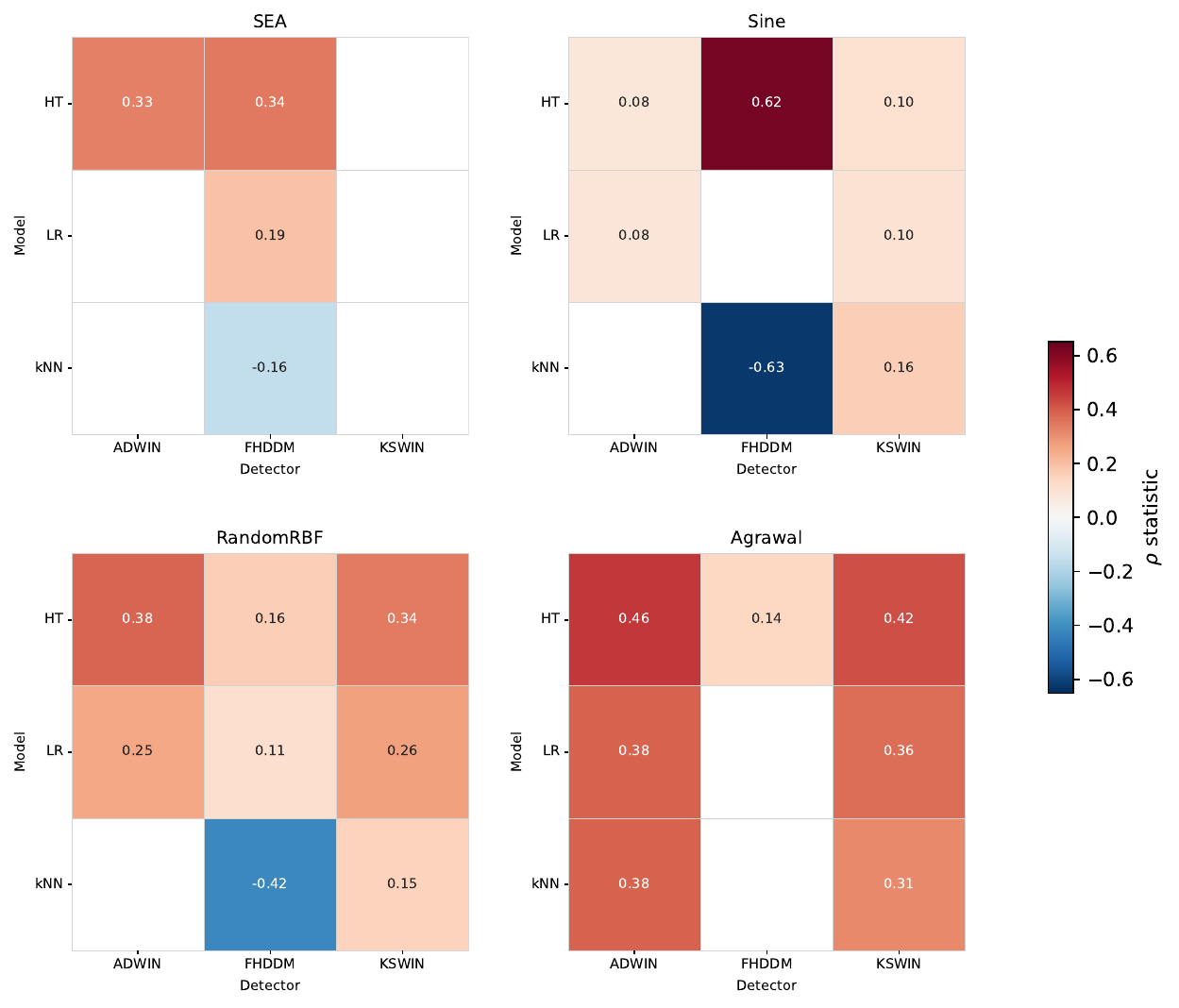}
    \caption{Results of Pearson correlation test; white fields indicate no statistically significant correlation, red indicates positive, blue indicates negative significant correlation}
    \label{fig:h3-stat}
\end{figure}

\paragraph{Hypothesis 3:} Timely detection is important\\
To investigate the effect of timely detection, in \cref{fig:h3}, we plotted the distance of the first detection against the ITTE. Next to plotting each experimental result with low alpha values, we averaged over the combinations of model and drift detector. For Sine (\cref{fig:h3-sine}) and SEA (\cref{fig:h3-sea}), the error rises significantly as the first detection moves further from the ground-truth drift event for all combinations apart from $k$NN and FHDDM, where we report a negative correlation between distance and ITTE. In contrast, for RandomRBF (\cref{fig:h3-randomrbf}), the error remains constant and even drops for the combination of $k$NN with the FHDDM detector. For Agrawal (\cref{fig:h3-agrawal}), we see that timely detection of a drift is beneficial within the first 100 samples, while afterwards there is no effect anymore.

We can again confirm these findings by a Pearson correlation test ($p<0.001$, see \cref{fig:h3-stat}), which yields a positive correlation (distance, ITTE) in 23 configurations, a negative correlation only for three combinations of $k$NN and FHDDM, and no significant results in 10 cases. 

We conclude that timely detection generally improves performance, with notable exceptions in specific model-detector combinations ($k$NN+FHDDM). The impact of detection delay is most pronounced in datasets with severe drift, where rapid adaptation is critical.

Interestingly, different drift detectors systematically yield vastly different detection times (most prominently visible in \cref{fig:h3-sine}). This aspect requires further investigation, relevant beyond the analysis of the relationship between drift detection and model performance in stream learning.

\begin{figure}[t]
    \centering
    \includegraphics[width=0.8\linewidth]{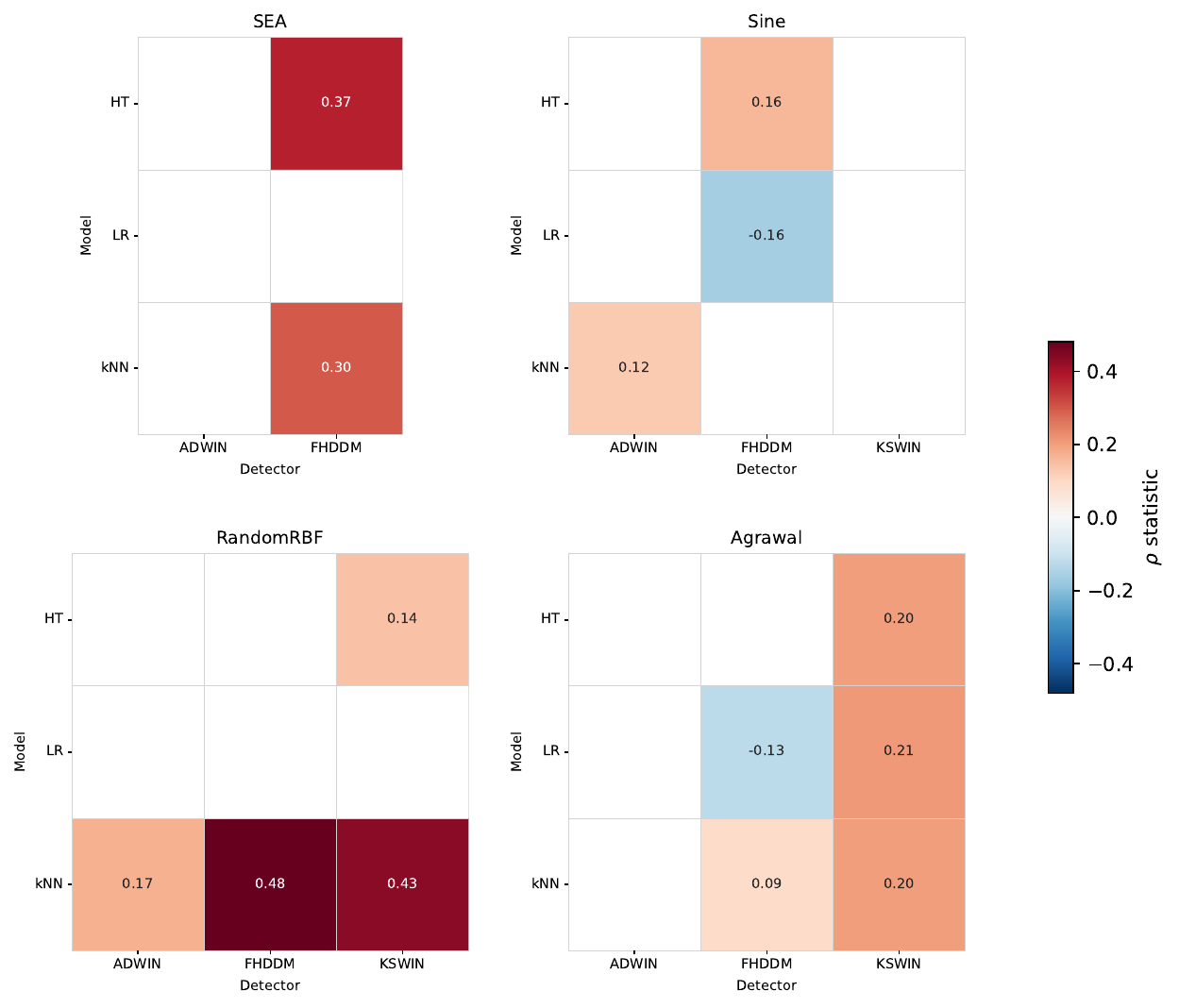}
    \caption{Results of Pearson correlation test; white fields indicate no statistically significant correlation, red indicates positive, blue indicates negative significant correlation}
    \label{fig:h4-stat}
\end{figure}

\paragraph{H4:} Too many correct alarms are harming model performance\\
To test this hypothesis, we investigated the correlation (Pearson, $p<0.001$) between the ITTE and the number of TPs.  For this analysis, we increased the acceptance window to 200 samples before and after the drift to capture a sufficient number of true positive events for statistical testing. The results are summarized in \cref{fig:h4-stat}. The test did not succeed for KSWIM on SEA due to too few FPs.

Significant correlations were found in 14/29 configurations, predominantly positive (12/14), linking more correct alarms to higher ITTE. Only two configurations (logistic regression with FHDDM) showed a negative correlation. The lack of significance in the remaining configurations is likely due to the low TP count. Thus, we provide evidence that excessive correct alarms can harm model performance, supporting H4.

\paragraph{Ablations}
As mentioned before, we also run all experiments without using a symmetric window to determine whether an alarm is considered a TP or FP. The results of these experiments can be found in the appendix. While we observe very small variations, these do not change any interpretation. The only hypothesis where we see a considerable effect is H4. Only considering the non-symmetric window yields considerably fewer TPs, which results in more configurations in which the test fails or only finds statistically non-significant correlations.

\section{Conclusion}
In this work, we empirically investigated the relationship between drift detector performance and the predictive accuracy of stream learning models. Contrary to common assumptions, we found that drift detection is not universally beneficial; its utility depends heavily on the severity of the drift. In scenarios with severe drift, timely detection significantly improves model performance. However, in settings with mild drift, the stability costs of false alarms and excessive updates often outweigh the benefits of adaptation, leading to degraded performance. We conclude that active adaptation strategies must be carefully calibrated to the specific drift characteristics of the data stream, as blindly prioritizing detector accuracy does not guarantee optimal model performance.

\begin{footnotesize}

\bibliographystyle{unsrt}
\bibliography{literature}

\end{footnotesize}

\appendix

\section{Results of Ablation}
\begin{figure}[h]
    \centering
        \begin{subfigure}[b]{0.48\textwidth}
        \includegraphics[width=\textwidth]{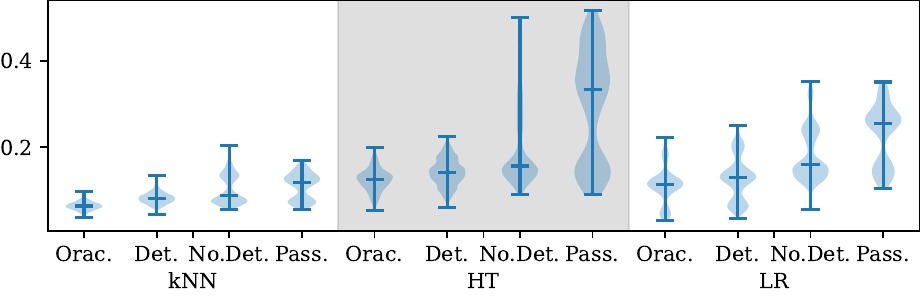}
        \caption{Sine}
    \end{subfigure}%
    \begin{subfigure}[b]{0.48\textwidth}
        \includegraphics[width=\textwidth]{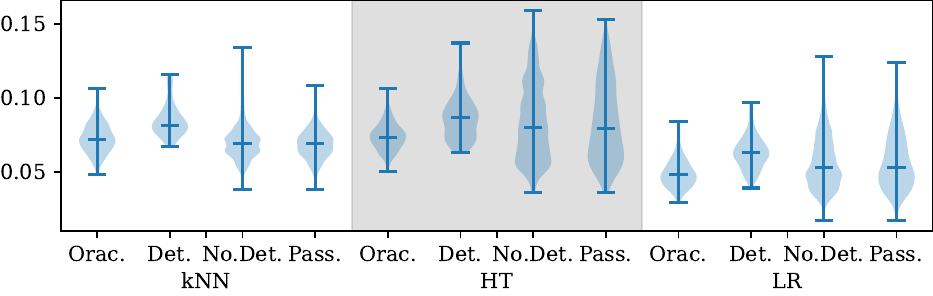}
        \caption{SEA}
    \end{subfigure}
    
    \begin{subfigure}[b]{0.48\textwidth}
        \includegraphics[width=\textwidth]{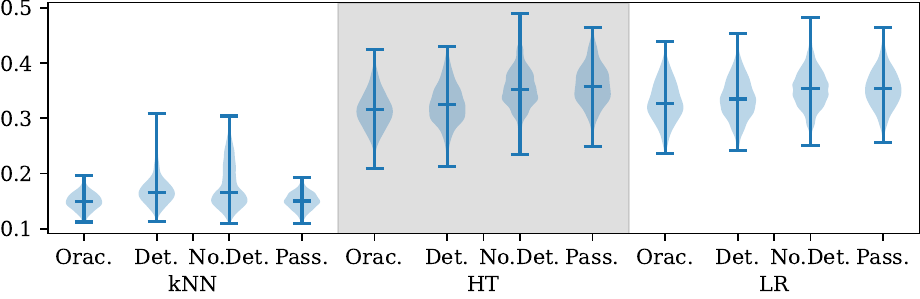}
        \caption{Random RBF}
    \end{subfigure}%
    \begin{subfigure}[b]{0.48\textwidth}
        \includegraphics[width=\textwidth]{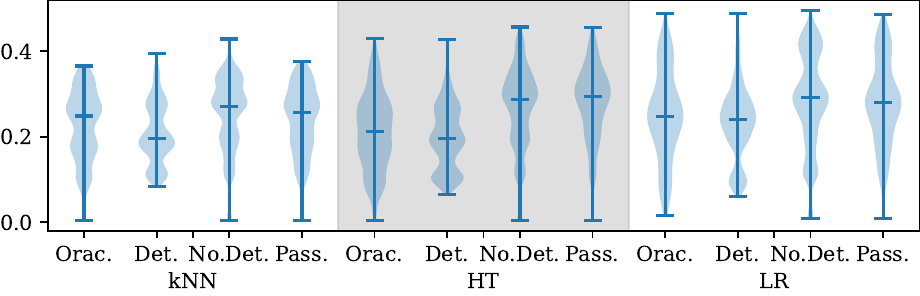}
        \caption{Agrawal}
    \end{subfigure}
    \caption{H1: non-symmetric window; ITTEs, grouped by ground oracle drift detector (Orac.), streams where the detector detected at least one true positive drift (Det.) and where it did not (No Det.), and passive approaches (Pass.).}
\end{figure}

\begin{figure}[h]
    \centering
        \begin{subfigure}[t]{0.45\textwidth}
        \includegraphics[width=\textwidth]{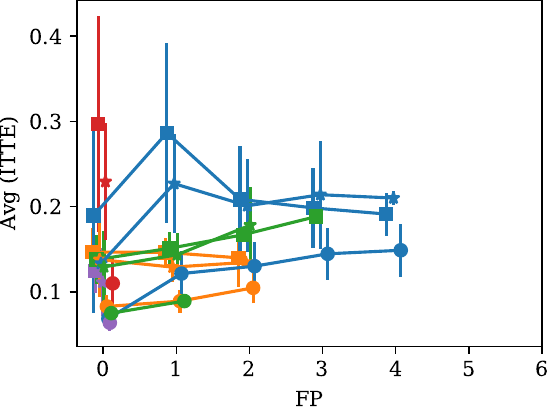}
        \caption{Sine}
    \end{subfigure}%
    \begin{subfigure}[t]{0.45\textwidth}
        \includegraphics[width=\textwidth]{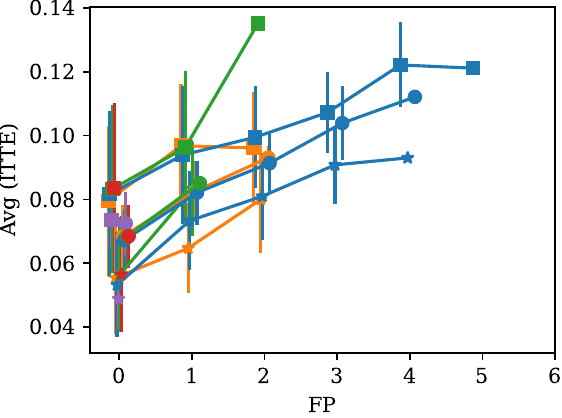}
        \caption{SEA}
    \end{subfigure}

    \begin{subfigure}[t]{0.45\textwidth}
        \includegraphics[width=\textwidth]{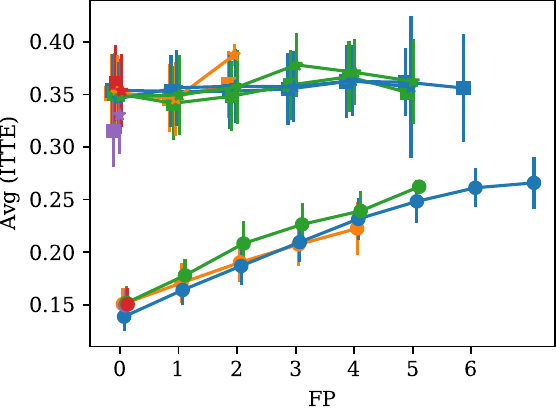}
        \caption{Random RBF}
    \end{subfigure}%
    \begin{subfigure}[t]{0.45\textwidth}
        \includegraphics[width=\textwidth]{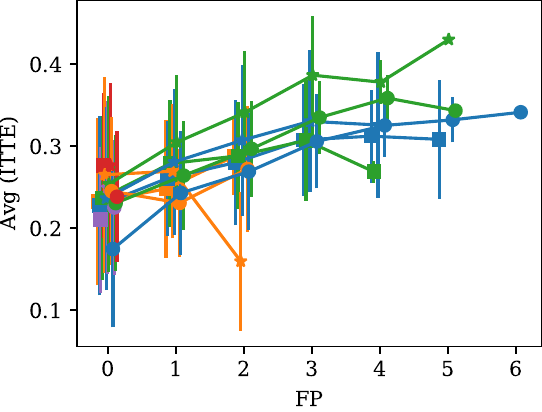}
        \caption{Agrawal}
    \end{subfigure}
    \begin{subfigure}[t]{\textwidth}
        \includegraphics[width=\textwidth]{imgs/H2/legend.pdf}
    \end{subfigure}
    \caption{H2: non-symmetric window; Average ITTE over increasing false positives for combinations of model and detection strategy}
\end{figure}

\begin{figure}[h]
    \centering
    \includegraphics[width=0.8\linewidth]{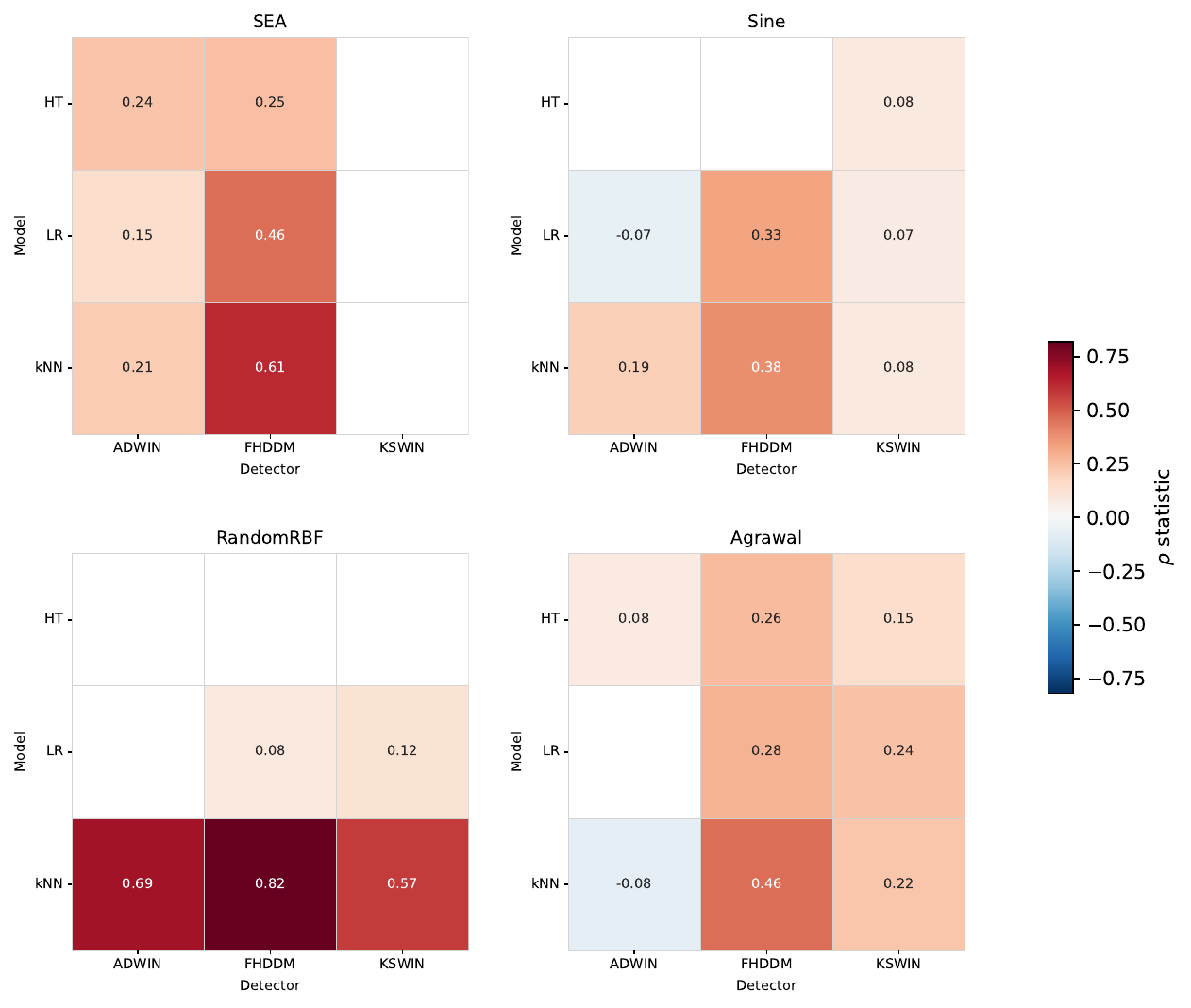}
    \caption{H2: non-symmetric window; Results of Pearson correlation test; white fields indicate no statistically significant correlation, red indicates positive, blue indicates negative significant correlation}
\end{figure}

\begin{figure}[t]
    \centering
        \begin{subfigure}[b]{0.48\textwidth}
        \includegraphics[width=\textwidth]{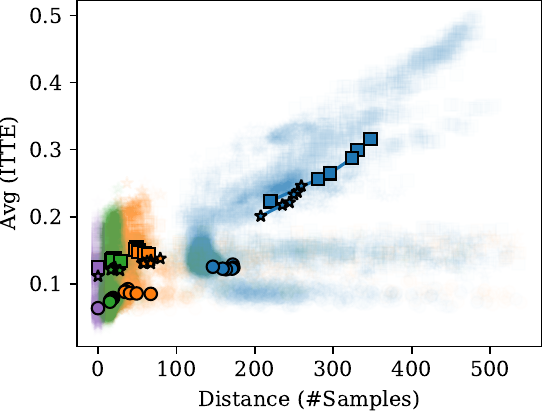}
        \caption{Sine}
    \end{subfigure}%
    \begin{subfigure}[b]{0.48\textwidth}
        \includegraphics[width=\textwidth]{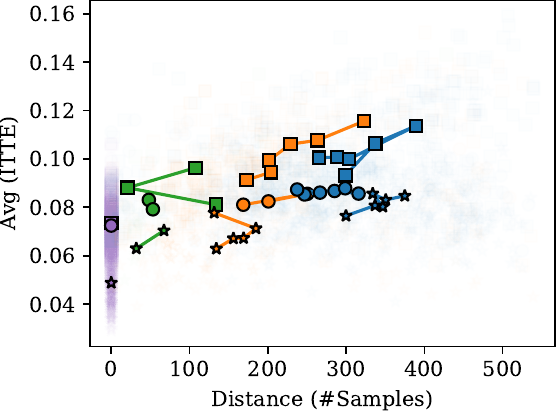}
        \caption{SEA}
    \end{subfigure}
    \begin{subfigure}[b]{0.48\textwidth}
        \includegraphics[width=\textwidth]{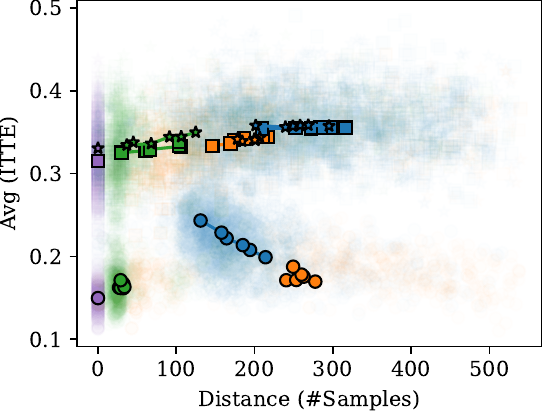}
        \caption{RandomRBF}
    \end{subfigure}%
    \begin{subfigure}[b]{0.48\textwidth}
        \includegraphics[width=\textwidth]{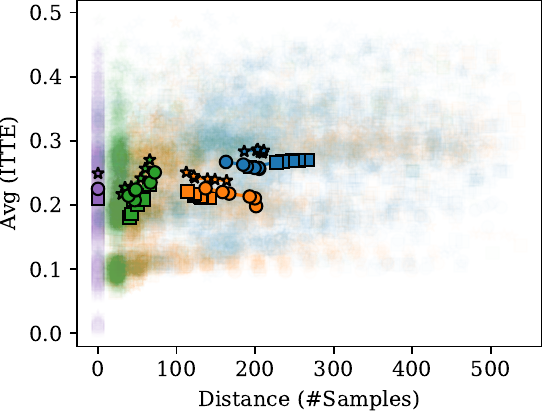}
        \caption{Agrawal}
    \end{subfigure}
    \begin{subfigure}[t]{\textwidth}
        \includegraphics[width=\textwidth]{imgs/H3/legend.pdf}
    \end{subfigure}
    \caption{H3: non-symmetric window; Average ITTE for varying distance of the first alarm for different combinations of model and detection strategy}
\end{figure}
\begin{figure}[t]
    \centering
    \includegraphics[width=0.8\linewidth]{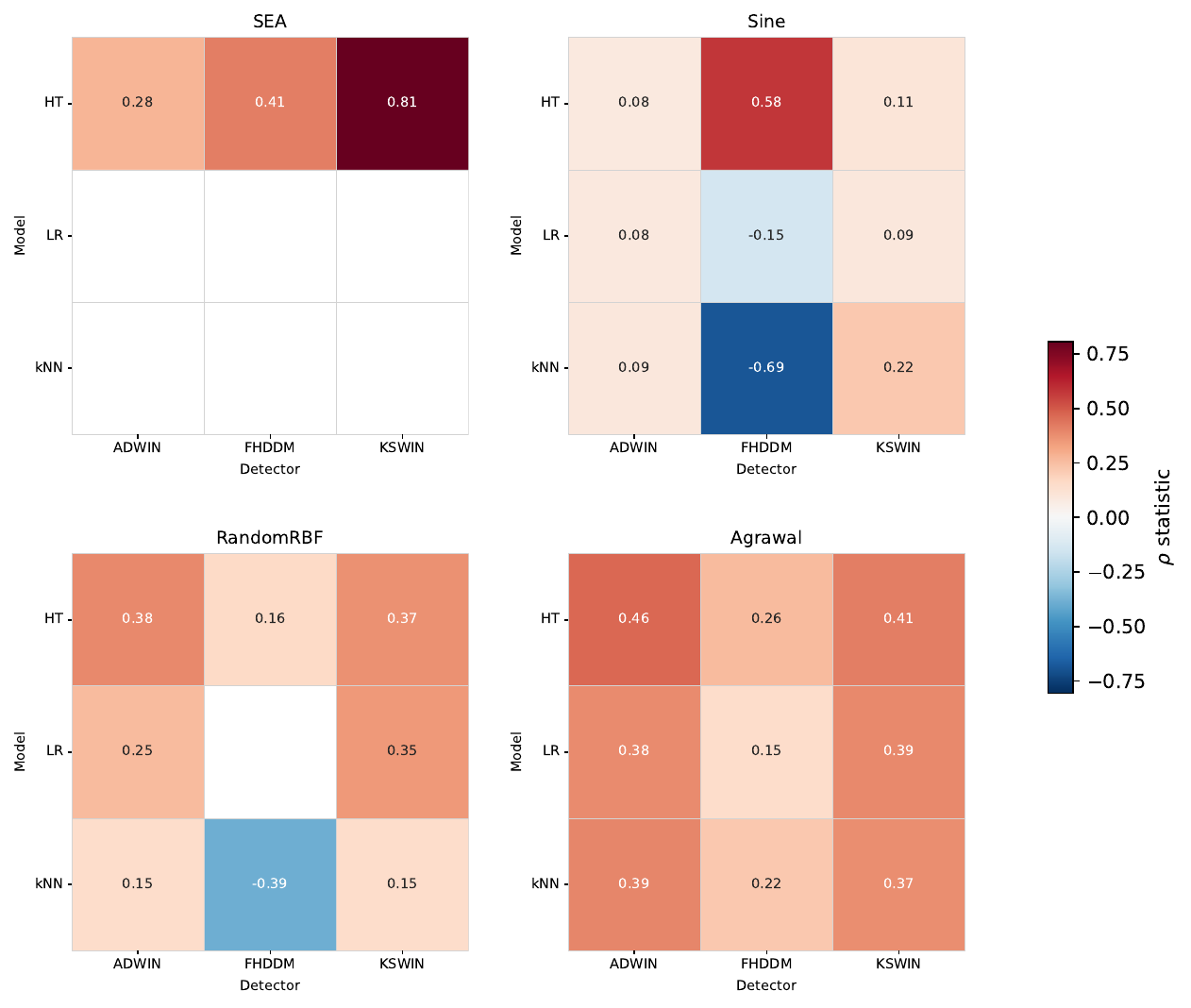}
    \caption{H3: non-symmetric window; Results of Pearson correlation test; white fields indicate no statistically significant correlation, red indicates positive, blue indicates negative significant correlation}
\end{figure}

\begin{figure}[t]
    \centering
    \includegraphics[width=0.8\linewidth]{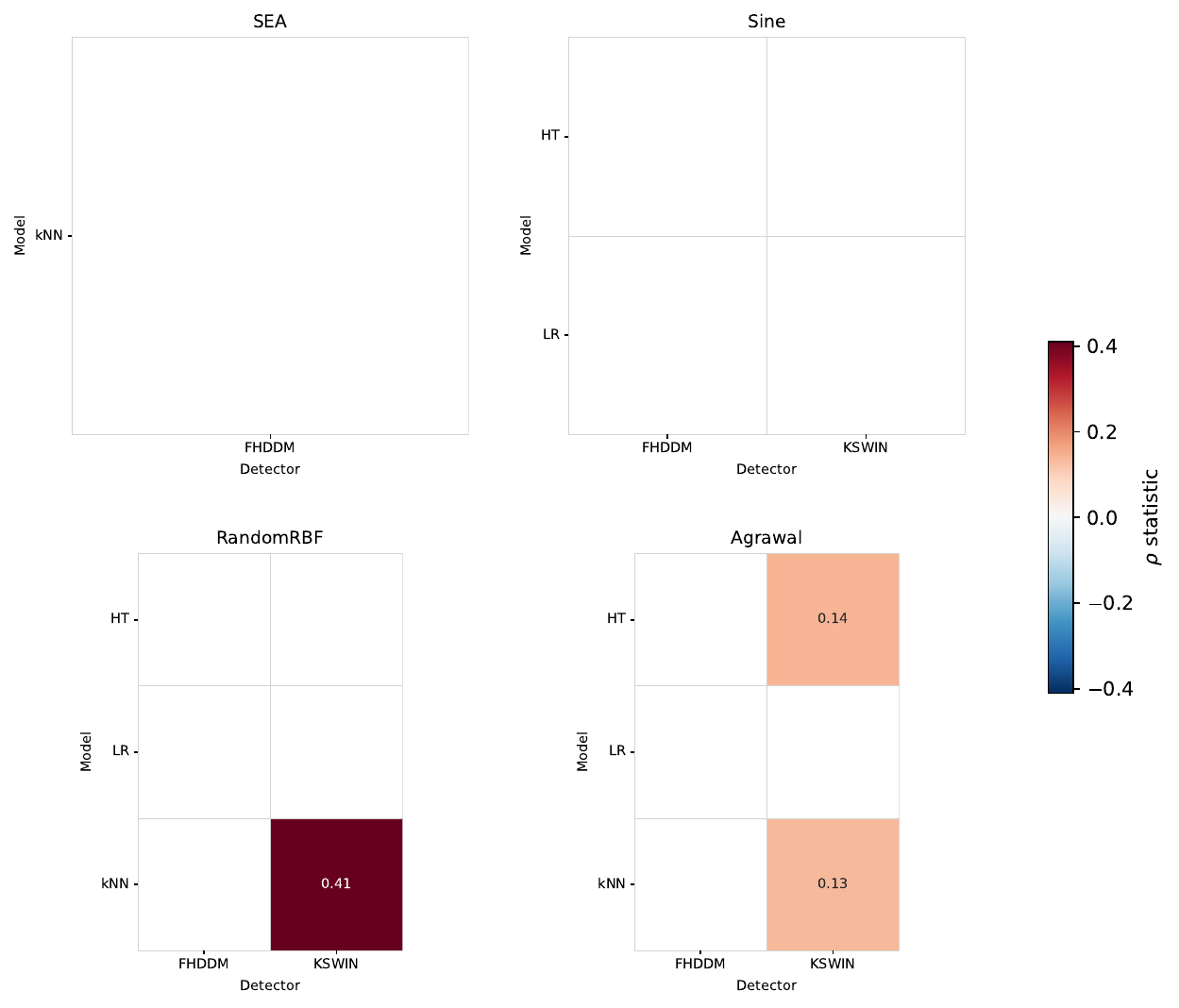}
    \caption{H4: non-symmetric window; Results of Pearson correlation test; white fields indicate no statistically significant correlation, red indicates positive, blue indicates negative significant correlation}
    \label{fig:h4-stat}
\end{figure}

\end{document}